\documentclass[journal,11pt,onecolumn,]{IEEEtran}
\usepackage{amsmath}
\usepackage{amsfonts}
\usepackage{amssymb}
\usepackage{algorithmic}
\usepackage{algorithm}
\usepackage{array}
\usepackage{tabularx}
\usepackage{array}
\usepackage{xcolor,colortbl}
\definecolor{Gray}{gray}{0.9}
\definecolor{Graay}{gray}{0.7}
\usepackage[font=normalsize]{subfig}
\usepackage{textcomp}
\usepackage{stfloats}
\usepackage{url}
\usepackage{float}
\usepackage{relsize}
\usepackage{multirow}
\usepackage{verbatim}
\usepackage{graphicx}
\usepackage{cite}
\newcommand{\RNum}[1]{\uppercase\expandafter{\romannumeral #1\relax}}

\usepackage{subfig,graphicx}
\usepackage[labelfont=bf]{caption}
\def\BibTeX{{\rm B\kern-.05em{\sc i\kern-.025em b}\kern-.08em
    T\kern-.1667em\lower.7ex\hbox{E}\kern-.125emX}}
\usepackage{balance}
\makeatletter
\newcommand{\thickhline}{%
    \noalign {\ifnum 0=`}\fi \hrule height 3pt
    \futurelet \reserved@a \@xhline
}
\makeatother

\begin{document}

\title{Deep Graph Clustering via Mutual Information Maximization and Mixture Model}

\author{Maedeh Ahmadi, Mehran Safayani, Abdolreza Mirzaei
        % <-this % stops a space
\thanks{The authors are with the Department of Electrical and Computer Engineering, Isfahan University of Technology, Isfahan
84156-83111, Iran (e-mail: m.ahmadi@ec.iut.ac.ir)}% <-this % stops a space
\thanks}

\maketitle

\begin{abstract}
Attributed graph clustering or community detection which learns to cluster the nodes of a graph is a challenging task in graph analysis. In this paper, we introduce a contrastive learning framework for learning clustering-friendly node embedding. Although graph contrastive learning has shown outstanding performance in self-supervised graph learning, using it for graph clustering is not well explored. We propose Gaussian mixture information maximization (GMIM) which utilizes a mutual information maximization approach for node embedding. Meanwhile, it assumes that the representation space follows a Mixture of Gaussians (MoG) distribution. The clustering part of our objective tries to fit a Gaussian distribution to each community. The node embedding is jointly optimized with the parameters of MoG in a unified framework. Experiments on real-world datasets demonstrate the effectiveness of our method in community detection. 
\end{abstract}


\section{Introduction}
\IEEEPARstart{G}{raphs} provide a way of representing a wide-variety of complex data in real-world systems. Several graph analysis approaches have emerged to extract useful information hidden in graphs. Community detection as an essential tool for graph analysis, has been applied to many real-world problems like social networks \cite{Yang2013}, citation networks \cite{Chen2010}, brain networks \cite{Nicolini2017} and Protein-Protein interaction (PPI) network \cite{Chen2006}.  Many community detection algorithms have been proposed; From shallow approaches \cite{Wang2017,Lesko2013} to deep ones \cite{arga,ijcai2019}. Several recent deep methods utilize graph convolutional network (GCN) \cite{kipf2016} to extract features from graphs \cite{mgae,AGC}. Many of these methods rely on local graph information (e.g. adjacency matrix reconstruction) which is not appropriate for graph clustering \cite{ijcai2019}.

Recently, contrastive approaches have demonstrated significant results in graph analysis tasks. For example, Deep Graph Infomax (DGI) \cite{DGI} have made great progress in node embedding via maximizing mutual information (MI) between node representations and graph summary. Despite the high performance of contrastive methods in node representation learning, they have not received sufficient attention for community detection. Obviously, better node embedding results in enhanced community detection.

In this paper, we aim to utilize the great potential of contrastive node embedding in community detection. One naïve approach is to apply a clustering algorithm, e.g. K-means, on the embedding result of a contrastive representation learning method to extract communities. This embedding space is suboptimal for the clustering task because the representation learning step is unaware of the downstream clustering task and is performed independent of it. To address this, we propose to learn a clustering-friendly node embedding which utilizes a contrastive method for node representation learning. As the contrastive embedding method, we follow the approach of \cite{DGI} which relies on mutual information maximization to learn representations. For clustering, we assume that the learnt node embedding space follows a Mixture of Gaussians (MoG) distribution. The parameters of the contrastive model and MoG are optimized in an iterative manner by gradient descent algorithm and Expectation Maximization (EM) \cite{EM}. Moreover, since many message passing algorithms are restricted to local messages, it is beneficial to employ a method which goes beyond direct neighbors to capture higher order information in the graph. To do so, we employ graph diffusion convolution (GDC) \cite{diffusion} which helps with the task of clustering by providing a global view of the graph.

Contributions of our method are summarized as follows: 1) we introduce a unified framework for learning clustering-friendly node embedding. 2) We utilize graph diffusion to benefit from the global view of the graph in the clustering task. 3) The experimental evaluations on real-world datasets demonstrate the effectiveness of our proposed method.

The rest of the paper is organized as follows. We review the related work of graph embedding and graph clustering in section  \RNum{2}. section \RNum{3} introduces a detailed description of the proposed method. Experimental results on two real-world datasets are presented section \RNum{4}. The conclusions are given in section \RNum{5}.

\section{Related works}
\subsection{Graph Embedding} 
Recently, approaches based on deep learning have made great progress in many fields of graph learning specially graph embedding. Early deep learning based researches were mostly rely on random walk objectives \cite{deepwalk,node2vec}. These methods take random walks along nodes and utilize neural language models (like SkipGram \cite{Le2014}) for node embedding. They assume that close nodes in the input graph, which co-occur in the same random sequence, should also be close in the embedding space. 
\IEEEpubidadjcol
Graph neural networks (GNNs) \cite{kipf2017, attention, Xu2019} have demonstrated strong representation power for attributed graph learning tasks. They follow a message passing mechanism to capture structural information of graph data. For unsupervised graph embedding, graph autoencoder-based methods \cite{arga, kipf2016,mgae} mainly try to reconstruct adjacency matrix so they impose closeness of fist-order neighbor nodes in the embedding space. Both of random walks and graph autoencoders-based methods over-emphasize the local proximity information \cite{DGI}. 

Recently, contrastive approaches have achieved state-of-the-art results in graph data analysis \cite{DGI,hassani,You2020, molaei}. They contrast samples from a desired distribution and another undesired one. Motivated by the excellent results of contrastive learning in visual representation learning \cite{Bachman2019,simclr}, graph contrastive algorithms propose to retain local and global structural information of graphs \cite{DGI,hassani}.

\subsection{Community Detection}
Many methods for detecting communities have been proposed. Early methods employ shallow approaches to community detection, mostly focusing on the information of network topology. Non-negative matrix factorization (NMF) \cite{Wang2017,Lesko2013} and laplacian eigenmaps \cite{Newman2006} are two widely used approaches in this area. Stochastic block model-based methods \cite{sbm} are also well explored. Modularity maximization is a popular goal to extract communities \cite{modularity}. To exploit both of content and structural information, several extended algorithms based on topic models \cite{topic} and NMF \cite{nmf,Wang2016} are proposed.  

As graph analysis problems and graph data get more complicated, deep learning based methods have demonstrated great performance in graph analysis tasks including community detection. As baseline methods among deep approaches, applying well-known clustering algorithms on embedding results of GAE and VGAE \cite{kipf2016} have better performance than many shallow algorithms. Some works present enhanced graph autoencoder-based methods with boosted results in graph clustering\cite{arga,AGC,mgae}. 

Recent methods try to combine clustering and graph embedding goals. \cite{ijcai2019} co-optimizes a graph attention-based reconstruction loss and the clustering loss of \cite{DEC}. \cite{Tsitsulin2020} maximizes modularity on the embedding space of a GCN. A probabilistic generative model which learns node embedding and community assignment jointly is proposed in \cite{vgraph}. In \cite{Shchur2019}, GCN is integrated with Bernoulli-Poisson probabilistic model \cite{Lesko2013} for overlapping community detection. \cite{relaxedkmeans} trains a graph auto-encoder to find an appropriate embedding space for relaxed K-means. A variational framework for learning clustering and node embedding is introduced in\cite{PR}.

\section{Method}
\subsection{Problem Formalization and Method Overview}
We consider community detection in attributed networks in this paper. The input is a graph $G=(V,E,X)$, where $V=(v_1,v_2, \ldots ,v_N )$  is the set of $N$ nodes and $E=\lbrace e_{ij} \rbrace$ is the edge set. $ X=\lbrace x_1; x_2; \ldots; x_N \rbrace$ are the attribute values where $x_i$  is the feature vector of node $v_i$. An adjacency matrix $A\in \mathbb{R}^{N\times N}$ encodes the structural connectivity of nodes where $A_{i,j}=1$ if  $(v_i, v_j )\in E$; otherwise $ A_{i,j}=0$.

The purpose of attributed community detection is to divide the nodes into K communities (or clusters) based on the attributes and structural information. 

Our proposed method considers the clustering and node embedding tasks in a joint manner. To achieve this goal, we assume that the node embedding space flows a Mixture of Gaussians distribution. We learn the parameters of the MoG and the contrastive method jointly. This results in a more cluster-friendly representation space which is more appropriate for K-means clustering algorithm to be applied to.

The proposed method includes two main parts: 1) Node embedding part which utilizes contrastive learning for extracting embedding vectors of the nodes. 2) Clustering part that tries to impose a Gaussian mixture distribution on the learned latent representation. The overall framework of our proposed method is shown in figure 1. 

\subsection{Node Embedding}
We use the contrastive framework of \cite{DGI} for learning node embedding on attributed networks. We maximize the mutual information between node representation vectors and a global graph summary vector. The objective is to train an encoder $\mathcal{E}$ such that $\mathcal{E}(X,A)=H=\lbrace h_1, h_2, \ldots, h_N  \rbrace \in \mathbb{R}^{N\times F}$
 represent node representations $h_i\in \mathbb{R}^F$  for each node $i$. We generate a negative graph $\tilde{G}$ by a corruption function $\tilde{G}=C(G)$ that shuffles the rows of $X$. The same encoder $\mathcal{E}$ is applied to the positive and negative graphs to obtain $H$ and $\tilde{H}$  representation matrices. Summary vector $s$ is obtained by the readout function $s=\mathcal{R}(H)=\sigma(1/N \sum_{i=1}^N h_i)$, with logistic sigmoid nonlinearity $\sigma$. Given representation vector $h$ and $s$, The following discriminator $\mathcal{D}$ distinguishes between representations from positive and negative graphs by assigning higher probabilities to representation vectors that the summary contain them:
 
 \begin{equation}
\label{eq:1}
\mathcal{D}(h, s)=\sigma(h^T Ws),
\end{equation}
where  $W$ is a learnable scoring matrix. To maximize the mutual information between $h_i$ and the summery vector $s$, the following cross-entropy loss is minimized:
\begin{equation}\label{eq:2}
\begin{array}{ll}
\mathcal{L}_{MI}=-\frac{1}{2N}\Big(\sum_{i=1}^N \mathbb{E}_{(X, A)}[\log \mathcal{D}(h_i, s)]
+\sum_{j=1}^N\mathbb{E}_{(\tilde{X}, \tilde{A})}[\log( 1-\mathcal{D}(\tilde{h}_j, s))]\Big).\end{array}
\end{equation}
The encoder is the following single-layer ${\rm GNC}$:
\begin{equation}\label{eq:3}
\mathcal{E}(X, A)={\rm PReLU}(\widehat{D}^{-\frac{1}{2}}\hat{A}\widehat{D}^{-\frac{1}{2}}X\Phi),
\end{equation}
where $\hat{A}=A+I_N$ is the adjacency matrix with self-connections and $\widehat{D}_{ii}=\sum_j A_{ij}$  is the corresponding degree matrix. $\Phi$ is a learnable transition matrix and ${\rm PReLU}$ represents parametric rectified linear unit function.

\begin{figure*}[!h]
\centering{
\includegraphics[width=6in]{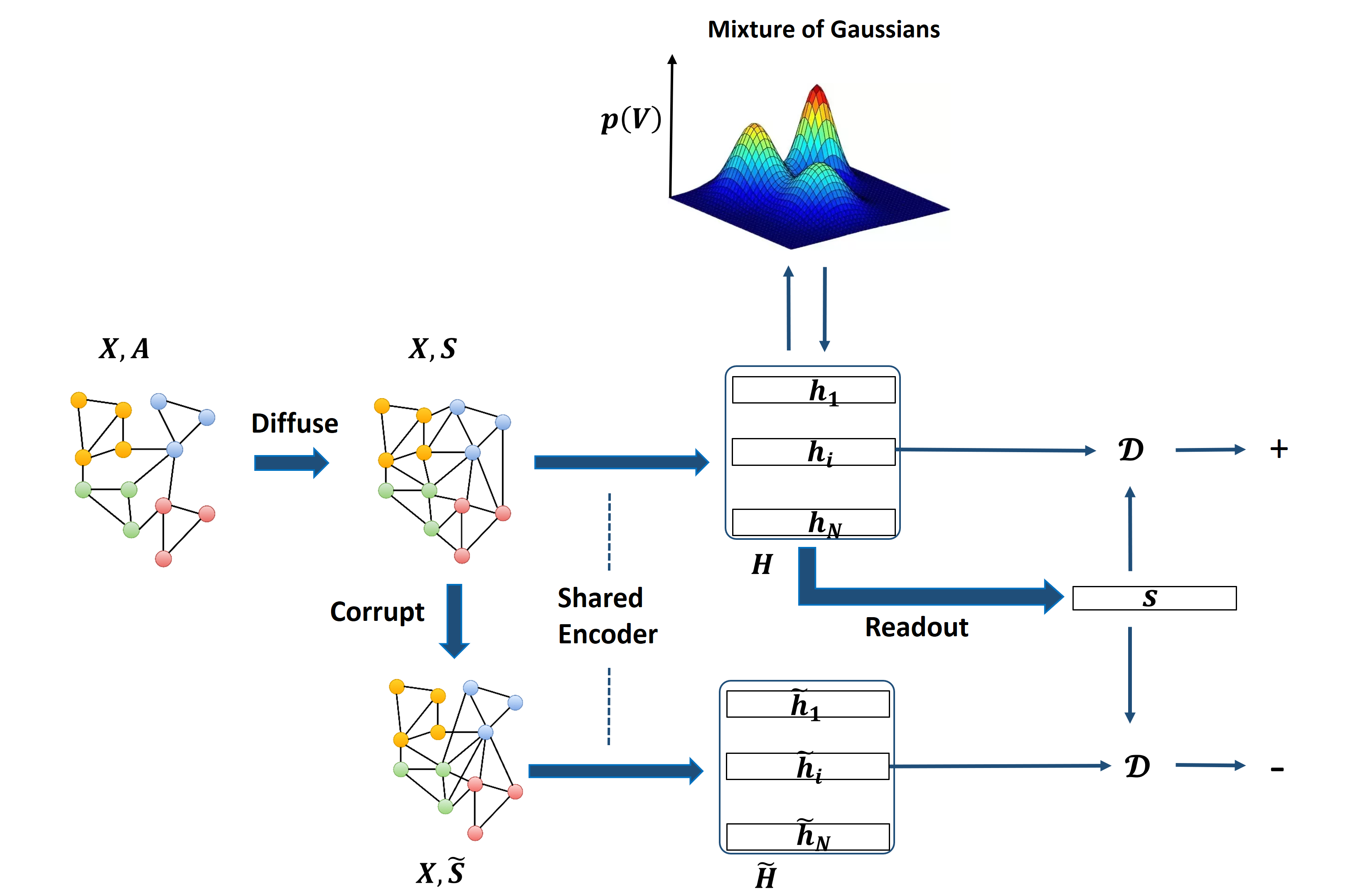}}
\caption{Our ${\rm GMIM}$ framework. A graph diffusion is applied to the graph. The resulted graph and its corrupted version are fed in to a shared encoder. The output features construct information maximization embedding objective. The clustering module (top) aims to enforce this representation to follow a ${\rm MoG}$ distribution. The embedding and clustering modules are trained jointly. }
\label{fig1}
\end{figure*}

\subsection{Graph Diffusion}
Message passing neural networks pass messages between immediate nodes of the graph. Although they try to aggregate the messages from higher-order neighbors in deep layers, most of them achieve their best performance with 2-layer networks because of over-smoothing phenomenon \cite{Li2018}. Limiting the messages of each layer to one-hop neighbors is restrictive and some methods try to capture higher-order information in the graph. One of the successful methods in this regard is Graph Diffusion Convolution (GDC) \cite{diffusion}. It replaces the adjacency matrix with a diffusion matrix which is formulated as:
\begin{equation}\label{eq:4}
S=\sum_{k=0}^{\infty}\Theta_k T^k,
\end{equation}
with generalized transition matrix $T$ and weighting coefficients $\Theta$. One popular example of graph diffusion is Personalized PageRank $({\rm PPR})$ \cite{pagerank}. Given adjacency matrix $A$ and related degree matrix $D$, $({\rm PPR})$ chooses $T=AD^{-1}$ and $\Theta_k=\alpha(1-\alpha)^k$ with teleport probability $\alpha\in [0,1]$. The closed-form solution for ${\rm PPR}$ diffusion is as below:
\begin{equation}\label{eq:5}
S^{\rm PPR}=\alpha(I_n-(1-\alpha) D^{-\frac{1}{2}} AD^{-\frac{1}{2}} )^{-1}.
\end{equation}
This diffusion matrix provides global view of a graph, acts as a low-pass filter and smooths out the neighborhood over the graph \cite{diffusion}. GDC can be integrated with any kind of graph-based model. In this work we utilize it as the input of our model instead of adjacency matrix.

\subsection{Gaussian Mixture Modeling for Community Detection}
Assume we have calculated a node embedding $h_i$ for every node $v_i$ of the graph by a node embedding model with parameters $\Psi$. We consider each node is generated from a multivariate Gaussian distribution. Then, the likelihood for all the nodes of the graph is a Gaussian mixture distribution:
\begin{equation}\label{eq:6}
p(V)=\prod_{i=1}^{|V|}\sum_{K=1}^K p(c_i=k)p(v_i |c_i=k;\Psi ,\mu_k, \Sigma_k ),
\end{equation}
here $c_i$ denotes the soft community assignment of node $i$ and $p(c_i=k)$  indicates the probability of node $i$ being assigned to community $k$. $p(v_i |c_i=k; \Psi, \mu_k, \Sigma_k )$  is a multivariate Gaussian distribution as follows:
\begin{equation}\label{eq:7}
p(v_i |c_i=k; \Psi, \mu_k, \Sigma_k )=N(h_i | \mu_k, \Sigma_k ).
\end{equation}
For simplicity of notations we denote $p(c_i=k)$  as $\pi_{i,k}$ where $\sum_{k=1}^K \pi_{i,k}=1$. So the parameters of the Gaussian mixture are $\Pi=\lbrace \pi_{i,k} \rbrace, M=\lbrace \mu_k \rbrace$  and $\sum=\lbrace \Sigma_k \rbrace$ for $i=1, \ldots,|V|$ and $k=1, \ldots, K$. We assume covariance matrices $\Sigma_k$ are diagonal. 

\subsection{Clustering-friendly Node Embedding}
We propose a clustering-promoting objective which outputs a latent space that is suitable for clustering. We assume that the learnt latent space follows a ${\rm MoG}$ distribution. Our defined objective function has two parts: embedding and clustering. The embedding part utilizes the self-learning objective of $\mathcal{L}_{MI}$   for node representation learning and the clustering module tries to enforce this representation to follow a ${\rm MoG}$ distribution. The later goal is achieved by minimizing the negative log-likelihood (${\rm NLL}$) under ${\rm MoG}$ distribution:
\begin{equation}\label{eq:8}
{L}_{NLL}=-\sum_{i=1}^{|V|}\log\sum_{k=1}^K\pi_{i,k}\mathcal{N}(h_i | \mu_k, \Sigma_k).
\end{equation}
Our total loss function is defined as:
\begin{equation}\label{eq:9}
\mathcal{L}=\omega\mathcal{L}_{MI}+\beta\mathcal{L}_{NLL},
\end{equation}
where $\mathcal{L}_{MI}$ and $\mathcal{L}_{NLL}$ are the mutual information loss and the negative log-likelihood (${\rm NLL}$) respectively. The weighs $\omega$ and $\beta$ balance between two terms of the objective function. After optimizing our objective, we have a K-means-friendly latent space on which we apply k-means algorithm to obtain the final clusters of nodes. 
\subsection{Inference}
The total loss function of (\ref{eq:9}) consists of two sets of parameters: Node embedding parameters $(\Psi)$ and  ${\rm MoG}$ parameters $(\Pi, M {\rm~ and}~ \Sigma)$. To optimize these parameters, we use an iterative approach by fixing one set and optimizing the other. We initialize the $\Psi$ parameters by training the model using (\ref{eq:2}) as the loss function. To initialize ${\rm MoG}$ parameters, we apply $K$-means algorithm on the achieved embedding from $\Psi$ initialization. We initialize $(\Pi, M, \Sigma)$ using the hard assignment results of $K$-means algorithm. The details of this iterative approach is described below.\\

\textbf{Fixing $\boldsymbol{\Psi}$ Parameters and Optimizing $\boldsymbol{(\Pi, M, \Sigma)}$}\\
Fixing deep network parameters, we use expectation maximization algorithm \cite{EM} to optimize $(\Pi, M, \Sigma)$. The following equations are used iteratively to update these parameters:
\begin{equation}\label{eq:10}
	\pi_{i,k}=\frac{N_k}{|V|},	
\end{equation}
%%%%%%%%%%%%
\begin{equation}\label{eq:11}
\mu_k=\frac{1}{N_k}  \sum_{i=1}^{|V|} \mathcal{V}_{ik}  h_i,
\end{equation}
%%%%%%%%%%%%%%%%%
\begin{equation}\label{eq:12}
\Sigma_k=\frac{1}{N_k}  \sum_{i=1}^{|V|} \mathcal{V}_{ik}  (h_i-\mu_k ) (h_i-\mu_k )^T,
\end{equation}
where
\begin{equation}\label{eq:13}
\mathcal{V}_{ik}=\frac{\pi_{i,k} \mathcal{N}(h_i |\mu_k, \Sigma_k )}{\sum_{k^{\prime}=1}^K \pi_{i,k^{'}} \mathcal{N}(h_i |\mu_{k^{'}}, \Sigma_{k^{'}} )},	
\end{equation}
and
\begin{equation}\label{eq:14}
	\mathcal{N}_k=\sum_{i=1}^{|V|}\mathcal{V}_ik   \quad                1\leq k \leq K.	
\end{equation}
More precisely, we update $\mathcal{V}_{ik}$ in $E$-step and $(\Pi, M, \Sigma)$ in the $M$-step of the ${\rm EM}$ algorithm.\\

\textbf{Fixing $\boldsymbol{(\Pi, M, \Sigma)}$ and Updating $\boldsymbol{\Psi}$ Parameters}\\
Fixing ${\rm MoG}$ parameters, we optimize the total loss function of (\ref{eq:6}) with respect to $\Psi$ parameters using stochastic gradient descent $({\rm SGD})$. $\Psi$ consists of the learnable scoring matrix $W$ of (\ref{eq:1}), encoder parameters $\Phi$ and ${\rm PReLU}$  parameters of  (\ref{eq:3}).
Our proposed method is summarized in Algorithm 1.

\section{Experiments}
\subsection{Benchmark Datasets}
We use two standard widely-used network datasets (Cora and Pubmed) for attributed graph community detection in our experiments. Cora and Pubmed are citation networks. Nodes represent papers and edges correspond to citations. Labels are papers topics and features are bag-of-word vectors. Features of Cora are binary vectors while Pubmed is represented by tf-idf weights. Table \ref{tab1} summarizes the detailed statistics of datasets.

\begin{table}
\renewcommand{\arraystretch}{1.2}
% \normalsize
\setlength{\tabcolsep}{11pt}
  \begin{center}
  \caption{Datasets statistics }
  \label{tab1}
    \begin{tabular}{c|c|ccc}
    \noalign{\hrule height 1pt}
         \textbf{Dataset}	&\textbf{Nodes}	&\textbf{Edges}	&\textbf{Features}	 
         &\textbf{Clusters}\\
    \noalign{\hrule height 1pt}  
         Cora &	2708 &	5429 &	1433 &	7\\
         Pubmed	& 19717	& 44338	& 500	& 3\\
    \noalign{\hrule height 1pt}     
    \end{tabular}
  \end{center}
\end{table}

\begin{algorithm}[H]
\caption{Gaussian Mixture Information Maximization.}
\begin{algorithmic}
\STATE \textbf{Require:}
\STATE \hspace{0.5cm}Graph $G=(V,E,X)$, number of clusters $K$, weight $\omega$, \STATE \hspace{0.5cm}${\rm PPR}$ parameter $\alpha$ and hidden dimension.
\STATE \textbf{Ensure:}
\STATE \hspace{0.5cm}Node embedding $H$, final community assignments.
\\
\STATE Calculate $S^{{\rm PPR}}$ by (\ref{eq:5})
\STATE Replace $A$ by $S^{{\rm PPR}}$ in (\ref{eq:2})
\STATE Initialize $H$ by optimizing (\ref{eq:2})
\STATE Initialize ${\rm MoG}$ parameters by applying $K$-means on $H$
\STATE {\bf for} $t=0$  to $T$ {\bf do}
\STATE \hspace{0.5cm}{\bf for} $t_1=0$  to $T_1$ {\bf do}
\STATE \hspace{1cm}Update $\Pi,M,\Sigma$ by equations (\ref{eq:10}), (\ref{eq:11}) and (\ref{eq:12})
\STATE \hspace{0.5cm} {\bf end }
\STATE \hspace{0.5cm} {\bf for} $t_2=0$ to $T_2$ {\bf do}
\STATE \hspace{1cm} Calculate mutual information loss by (\ref{eq:2})
\STATE \hspace{1cm} Calculate negative log-likelihood loss by (\ref{eq:8})
\STATE \hspace{1cm} Update $\Psi$ parameters by SGD on (\ref{eq:9})
\STATE \hspace{0.5cm}   {\bf end }
\STATE {\bf end }
\STATE Get the final community assignments by applying $K$-means on $H$.
\end{algorithmic}
\label{alg1}
\end{algorithm}
\subsection{Baseline Methods}
We compare GMIM with the following baseline methods. These approaches are categorized into three groups:
\begin{enumerate}
\item Methods which use node features only: K-means and spectral clustering \cite{Ng2002} are two common clustering methods. Spectral-F is a spectral clustering method which considers the cosine similarity between node features as the similarity matrix. 
\item Methods which use graph structure only: Spectral-G considers the adjacency matrix as the similarity matrix. DeepWalk \cite{deepwalk} generates random paths along a graph and use them to train SkipGram language model to learn node embedding. GraphEncoder \cite{tian} trains a stacked sparse auto-encoder to obtain node embedding. DNGR \cite{Cao2016} uses stacked denoising autoencoders to learn each node representation. K-means in applied to the learnt latent space of the three later methods. vGraph \cite{vgraph} is a probabilistic generative model which performs graph clustering and node embedding jointly.
\item Methods which use both node features and graph structure: TADW \cite{richtext} adds node features to DeepWalk framework. GAE and VGAE \cite{kipf2016} integrate (variational) autoencoder and graph neural networks for node embedding. MGAE \cite{mgae} introduces a marginalized graph autoencoder for graph clustering. ARGA and ARVGA \cite{arga} use an adversarial training scheme to impose a prior distribution on latent space of GAE and VGAE. DGVAE \cite{dirichlet} presents a graph variational generative model which uses the Dirichlet distributions as priors on the latent variables. AGC \cite{AGC} designs a high-order graph convolution to take smooth node features for enhancing clustering results. CommDGI \cite{commdgi} incorporates contrastive learning to learn cluster assignment of the nodes. DAEGC \cite{ijcai2019} optimizes graph reconstruction loss and a clustering loss jointly. SENet \cite{senet} uses a spectral clustering loss to learn node embeddings. GC-VGE \cite{PR} introduces a joint framework for clustering and representation learning by utilizing a variational graph embedding mechanism. EGAE \cite{relaxedkmeans} learns graph auto-encoder and relaxed K-means simultaneously to obtain a latent space which is ideal for relaxed K-means. This method has not reported its results for Pubmed dataset. GALA \cite{gala} proposes a symmetric graph convolutional autoencoder in which the encoder and decoder are based on Laplacian smoothing and sharpening respectively. DBGAN \cite{cvpr} introduces an adversarial framework to learn node embeddings.
\end{enumerate}

\subsection{Evaluations Metrics and Implementation Details}
We report five evaluation metrics to measure the performance of graph clustering: clustering accuracy (ACC), normalized mutual information (NMI), adjusted rand index (ARI), F-Score (F1) and precision (P). The higher values of all these metrics indicates the better results. We run our algorithm 10 times on each dataset and report the average of obtained metrics. 
For the encoder, we set the size of hidden dimension to 512 and 256 for Cora and Pubmed respectively. Due to memory limitations we could not increase the hidden size for Pubmed. We set $\alpha=0.2$ for ${\rm PPR}$ diffusion. The weight $\omega$ is set to $25000$ and $1000$ for Cora and Pubmed respectively, to balance two terms of objective function. At the start of training we set $\beta$ to zero and as training progresses, we gradually increase it to reach one. We use the Adam ${\rm SGD}$ optimizer with learning rate of $0.001$ in both initialization and training phases for both datasets. $T_1$ and $T_2$ are set to one for both datasets. We train the model for $200$ and $1000$ epochs on Cora and Pubmed respectively.

\subsection{Experimental Results}
Our experimental results are summarized in tables \ref{tab2} and \ref{tab3}. F, G and F\&G indicate the methods which use only node features, graph structure or both of features and structure information, respectively. As shown in these tables, methods using both feature and structure generally outperform the methods using only one source of information. This indicates the importance of these information for clustering task.
Our method has superior performance to TADW. This method does not utilize GNN for node embedding. Our method significantly outperforms GAE, VGAE, MGAE, ARGA and ARVGA. These are two-stage methods which perform node embedding and clustering stage independently. DAEGC has a unified framework for clustering and representation learning but its objective rely on adjacency matrix reconstruction which over-emphasize proximity information. AGC captures the global structure of a graph by utilizing high-order graph convolution. This is similar to how our framework benefits from graph diffusion which gives a global view of the graph. However, this method performs representation learning and clustering independently. The proposed method has higher performance than GALA and DBGAN. These methods learn cluster assignment and node embeddings in two separate stages. CommDGI, GC-VGE and EGAE learn cluster assignments and node representations jointly but they have lower performance in comparison to our method. 

\setlength{\tabcolsep}{10pt}
\begin{table*}[!h]
\renewcommand{\arraystretch}{1}
\normalsize
\begin{center}
\caption{Clustering results on Cora dataset }
\label{tab2}
\begin{tabular}{ p{3.3cm} | p{1.2cm}| p{1.2cm} p{1.2cm} p{1.2cm} p{1.2cm} p{1.2cm}}
\noalign{\hrule height 1pt}
\textbf{Method}	& \textbf{Info.}	& \textbf{ACC}	& \textbf{NMI}	& \textbf{ARI}	& \textbf{F1}	& \textbf{P}\\
\noalign{\hrule height 1pt}
$K$-means &	F &	49.2	& 32.1 & 	23.0 & 	36.8 &	36.9\\
Spectral-F \cite{Ng2002}
& F	& 34.7 & 	14.7 & 	7.1 & 	- & 	-\\\hline
Spectral-G \cite{Ng2002}
& G	& 31.46	& 9.69 & 	0.35 &	29.67 &	18.07\\
DeepWalk \cite{deepwalk}
& G &	56.20 &	39.87 &	32.18 &	47.6 &	5.48\\
GraphEncoder \cite{tian}
& G &	32.5 &	10.9 & 	0.6 & 	29.8 & 	18.2\\
DNGR \cite{Cao2016}
& G	 & 44.39 &	33.31 &	15.86 &	34.68 & 	27.86\\
vGraph \cite{vgraph}
& G	& 28.7 &	34.5 &	31.2 &	30.5 &	-\\\hline
TADW \cite{richtext}
&F$\&$G	&55.00	&36.59	&26.40	&41.52	&36.50\\
GAE \cite{kipf2016}
&F$\&$G	&60.34	&44.85	&36.73	&58.72	&61.39\\
VGAE \cite{kipf2016}
&F$\&$G	&63.56	&47.45	&39.42	&63.75	&65.64\\
MGAE \cite{mgae}
&F$\&$G	&63.43	&45.57	&38.01	&38.01	&-\\
ARGA \cite{arga}
&F$\&$G	&60.84	&42.21	&36.88	&60.49	&63.38\\
ARVGA \cite{arga}
&F$\&$G	&62.83	&45.93	&38.00	&63.17	&64.80\\
DGVAE \cite{dirichlet}
&F$\&$G	&64.42	&47.64	&38.42	&62.69	&64.90\\
AGC \cite{AGC}
&F$\&$G	&68.92	&53.68	&48.6	&65.61	&-\\
CommDGI \cite{commdgi}
&F$\&$G	&69.8	&57.9	&50.2	&68.4	&-\\
DAEGC \cite{ijcai2019}
&F$\&$G	&70.4	&52.8	&49.6	&68.2	&-\\
SENet \cite{senet}
&F$\&$G	&71.92	&55.08	&48.96	&-	&-\\
GC-VGE \cite{PR}
&F$\&$G	&70.67	&53.57	&48.15	&69.48	&70.51\\
EGAE \cite{relaxedkmeans}
&F$\&$G	&72.42	&53.96	&47.22	&-	&-\\
GALA \cite{gala}
&F$\&$G	&74.6	&57.7	&53.2	&-	&-\\
DBGAN \cite{cvpr}
&F$\&$G	&74.8	&56.0	&54.0	&-	&-\\
\noalign{\hrule height 1pt}
 GMIM (ours)&	F$\&$G	& 75.65	& 60.30	& 54.74	& 71.20	& 78.13\\
\noalign{\hrule height 1pt}
\end{tabular}
\end{center}
\end{table*}

% Table 3
\setlength{\tabcolsep}{10pt}
\begin{table*}[!h]
\renewcommand{\arraystretch}{1}
\normalsize
\begin{center}
\caption{Clustering results on Pubmed dataset }
\label{tab3}
\begin{tabular}{ p{3.3cm} | p{1.2cm}| p{1.2cm} p{1.2cm} p{1.2cm} p{1.2cm} p{1.2cm}}
\noalign{\hrule height 1pt}
\textbf{Method}	& \textbf{Info.}	& \textbf{ACC}	& \textbf{NMI}	& \textbf{ARI}	& \textbf{F1}	& \textbf{P}\\
\noalign{\hrule height 1pt}
$K$-means	& F	& 55.59	& 24.34	& 21.54	& 56.04	& 46.08\\
Spectral-F \cite{Ng2002}
&F	& 60.20	& 30.90	& 27.7	& -	& -\\\hline
Spectral-G \cite{Ng2002}
& G	& 37.98	& 10.30	& 26.67	& 50.54	& -0.02\\
DeepWalk \cite{deepwalk}
& G	& 64.98	& 26.44	& 27.42	& 63.46	& 65.24\\
GraphEncoder \cite{tian}
& G	& 53.1	& 20.9	& 18.4	& 50.6	& 45.6\\
DNGR \cite{Cao2016}
& G	& 25.53	& 20.11	& 8.29	& 15.57	& 19.26\\
vGraph \cite{vgraph}
& G	& 26.00	& 22.40	& 18.50	& 33.20	& -\\\hline
TADW \cite{richtext}
&F$\&$G	&46.82	&9.47	&5.78	&51.22	&38.34\\
GAE \cite{kipf2016}
&F$\&$G	&64.43	&24.85	&23.57	&64.07	&65.26\\
VGAE \cite{kipf2016}
&F$\&$G	&64.67	&23.94	&23.41	&64.77	&64.53\\
MGAE \cite{mgae}
&F$\&$G	&43.88	&8.16	&3.98	&41.98	&-\\
ARGA \cite{arga}
&F$\&$G	&65.07	&29.23	&26.79	&64.11	&69.27\\
ARVGA \cite{arga}
&F$\&$G	&62.01	&26.62	&22.46	&61.66	&68.41\\
DGVAE \cite{dirichlet}
&F$\&$G	&67.56	&28.72	&24.92	&64.35	&67.10\\
AGC \cite{AGC}
&F$\&$G	&69.78	&31.59	&31.19	&68.72	&-\\
CommDGI \cite{commdgi}
&F$\&$G	&69.90	&35.70	&29.2	&69.2	&-\\
DAEGC \cite{ijcai2019}
&F$\&$G	&67.10	&26.60	&27.8	&65.9	&-\\
SENet \cite{senet}
&F$\&$G	&67.59	&30.61	&29.66	&-	&-\\
GC-VGE \cite{PR}
& F$\&$G	&68.18	&29.70	&29.76	&66.87	&69.39\\
GALA \cite{gala}
&F$\&$G	&69.39	&32.73	&32.1	&-	&-\\
DBGAN \cite{cvpr}
&F$\&$G	&69.40	&32.40	&32.7	&-	&-\\
\noalign{\hrule height 1pt}
 GMIM (ours)&	F$\&$G	& 70.87	& 32.43	& 33.25	& 69.19	& 70.83\\
\noalign{\hrule height 1pt}
\end{tabular}
\end{center}
\end{table*}

\subsection{Ablation Study}
To study the effect of using diffusion graph, we train the model using only the mutual information loss ($\mathcal{L}_{MI}$) without replacing the adjacency matrix with the diffusion graph. We apply K-means and Gaussian mixture model (GMM) on the learnt embedding and the average result of 10 runs is reported in table \ref{tab4} as MI+K-means/GMM. We then replace the adjacency matrix with diffusion matrix and repeat the above experience. MI+Diffusion+K-means/GMM shows the related results. Comparing MI+K-mean/GMM with MI+Diffusion+K-means/GMM shows that the diffusion can improve the clustering performance.
To evaluate the effectiveness of our unified framework for learning clustering-friendly node embedding, we compare it against MI+Diffusion+K-means/GMM. The fact that our method outperforms DGI+Diffusion+K-means/GMM confirms the effectiveness of jointly optimizing MI and NLL objectives.
\subsection{Visualization}
\begin{figure*}[htp]
\centering
\subfloat[(a) Raw features]{\includegraphics[width=2in]{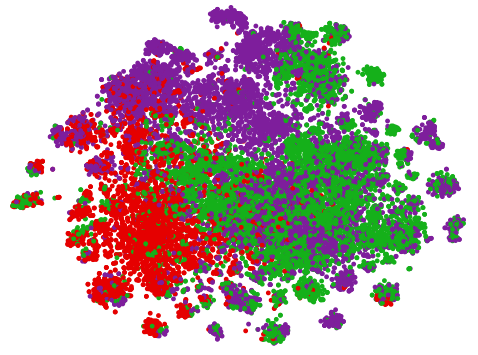}%
\label{fig_2_1}}
\subfloat[(b) MI+Diffusion]{\includegraphics[width=2in]{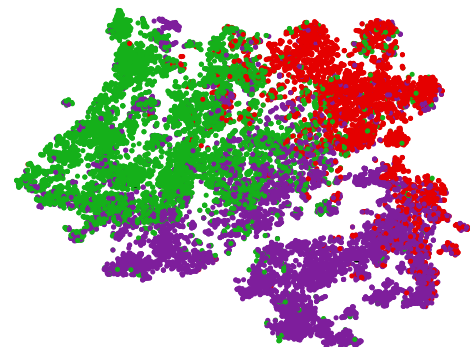}%
\label{fig_2_2}}
\subfloat[(c) GMIM]{\includegraphics[width=2in]{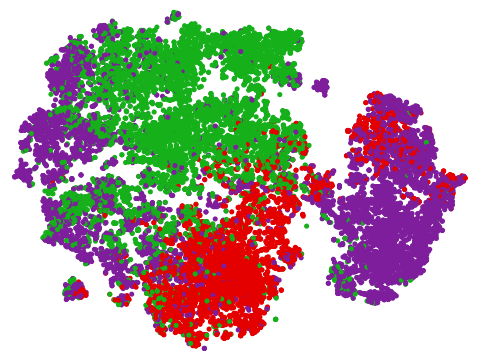}%
\label{fig_2_3}}
\caption{$2D$ visualization of node embeddings on Pubmed dataset. Different classes are shown by different colors.}
\label{fig2}
\end{figure*}

We visualize the node representations of Pubmed dataset in a two-dimensional space using t-SNE \cite{tsne}. The result is shown in Fig. \ref{fig2}. The latent space of MI+Diffusion and GMIM are shown in Fig. \ref{fig_2_2} and \ref{fig_2_3} respectively. Although both of these subfigures show more separable embedding in comparison to raw features, the latent space of GMIM has better fit to a mixture of Gaussians distribution. So, it is more clustering-friendly and is more appropriate for K-means to be applied to.

\setlength{\tabcolsep}{14pt}
\begin{table*}[!h]
\renewcommand{\arraystretch}{1.1}
\normalsize
\begin{center}
\caption{Ablation study}
\label{tab4}
\begin{tabular}{ p{4.5cm} | p{1cm} | p{1.2cm} p{1.2cm}|p{1.2cm}p{1.2cm}}
\hline
\noalign{\hrule height 1pt}
\multirow{2}{*}{\textbf{Method}} &  \multirow{2}{*}{\textbf{Input}} & \multicolumn{2}{c|}{\textbf{Cora}} & \multicolumn{2}{c}{\textbf{Pubmed}}\\
 & & ACC	& NMI & 	ACC	& NMI \\
\hline
MI+K-means &	$X$,$A$	& 72.15	& 56.50	& 66.37	& 31.08\\
MI+GMM	& $X$,$A$	& 67.69	& 53.52	& 64.36	& 28.04\\\rowcolor{Gray}
MI+Diffusion+K-means &	$X$,$S$	& 74.45	& 58.99	& 67.87	& 31.40\\\rowcolor{Gray}
MI+Diffusion+GMM	& $X$,$S$	& 69.47	& 56.32	& 65.85	& 26.36\\\rowcolor{Graay}
GMIM &	$X$,$S$	& 75.65	& 60.30	& 70.87	& 32.43\\ 
\hline
\noalign{\hrule height 1pt}
\end{tabular}
\end{center}
\end{table*}

\section{Conclusions}
In this paper we introduce a clustering-promoting objective for node embedding. Our proposed method utilizes contrastive learning to produce a clustering-friendly latent space by assuming that the learnt representation follows a mixture of Gaussians distribution. The embedding and clustering objectives are optimized in a unified framework to benefit each other. Our experiments show that incorporating the clustering-directed objective function can enhance the clustering ability of graph contrastive learning. We evaluated the proposed method on real-world datasets to show its effectiveness Empirical results demonstrate the favorable performance of our method compared with state-of-the-art methods.

\bibliographystyle{IEEEtran}
\bibliography{Arxiv_paper}

\end{document}